\documentclass{styles/svproc}
\usepackage[utf8]{inputenc}
\usepackage{tikz}
\usepackage{multirow}
\usepackage{amsmath}
\usepackage{amsfonts}
\usepackage{amssymb}
\usepackage{soul}
\usepackage{comment}
\usepackage[ruled,vlined]{algorithm2e}
\newcommand{\largehline}{\specialrule{.16em}{.3em}{.3em}}
\newcommand{\mediumhline}{\specialrule{.10em}{.2em}{.2em}}
\newcommand{\ttitle}[1]{#1}
\usepackage{ctable}
\usepackage{listings}
\usepackage{diagbox}
\usepackage{arydshln}
\title{Does Structure Matter? Leveraging Data-to-Text Generation for Answering Complex Information Needs}

\author{Hanane Djeddal$^*$, Thomas Gerald$^*$, Laure Soulier$^*$, \\Karen Pinel-Sauvagnat$^{**}$, Lynda Tamine$^{**}$}
\institute{\small
    $^*$ Sorbonne Université, CNRS, LIP6, F-75005 Paris, France\\ 
    $^{**}$ Université Paul Sabatier, IRIT, Toulouse, France \\ 
    \texttt{
      \{hanane.djeddal, thomas.gerald, laure.soulier\}@lip6.fr \\ \{sauvagnat, tamine\}@irit.fr \\ }}
\usepackage{enumitem}

\setlist[itemize]{label=\textbullet}
\begin{document}

\maketitle

\begin{abstract}
In this work, our aim is to provide a structured answer in natural language to a complex information need. Particularly, we envision using generative models from the perspective of data-to-text generation. We propose the use of a content selection and planning pipeline which aims at structuring the answer by generating intermediate plans.
The experimental evaluation is performed using the TREC Complex Answer Retrieval (CAR)  dataset. 
We evaluate both the generated answer and its corresponding structure and show the effectiveness of planning-based models in comparison to a text-to-text model.
\vspace{-0.6cm}
\end{abstract}

\begin{figure}[t]
    \centering
    \includegraphics[width=0.96\textwidth]{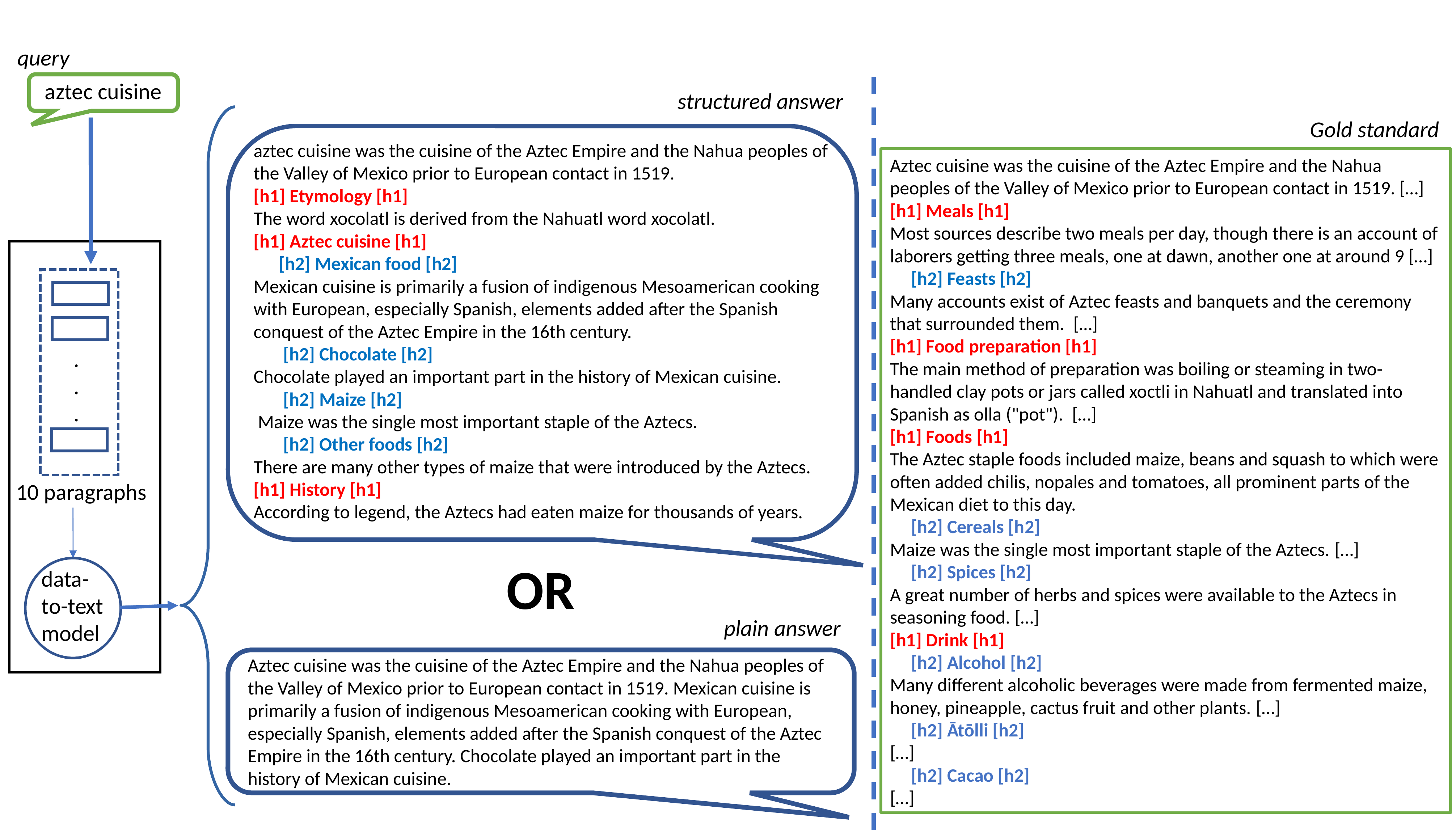}
    \vspace{-0.4cm}
    \caption{Example of a query from the CAR dataset \cite{dietz2017trec} and  variants of outputs (structured or plain answers) obtained using a sequential DTT planning-based model.}
    \vspace{-0.6cm}
    \label{fig:Example}
\end{figure}
\vspace{-0.4cm}
\section{Introduction}
\vspace{-0.2cm}

Complex search tasks (e.g., exploratory) involve open-ended and multifaceted queries that require information retrieval (IR) systems to aggregate  information over 
multiple unstructured documents \cite{white2009exploratory}.
To address these requirements, most
interactive IR methods adopt the dynamic multi-turn retrieval approach by designing session-based predictive models relying on Markov models \cite{YangGZ15}, query-flow graphs \cite{Hassan14} for relevance prediction and sequence-to-sequence models for query suggestion \cite{Sordoni14,Wasi19}.  
One drawback of those approaches remains in the iterative querying process, requesting users to visit different contents to complete their information need.
Moreover, while there is a gradual shift today towards new interaction paradigms through natural-sounding answers \cite{Culpepper18,Tan2018SNetFA}, most approaches still rely on a ranked list of documents as the main form of answer.

\indent In this paper, we envision solving complex search tasks triggered by open-ended queries, by considering single-turn (vs. multi-turn) interaction with users and providing  natural language generated answers (vs. a ranked list of documents).
We focus on the upstream part of the search process, once a ranked list of candidate documents has been identified in response to a complex information need. In a close line of research, 
open-domain QA attempt to retrieve and reason over multiple seed passages either to extract  \cite{Asai2020Learning,Chen2017ReadingWT}  or to generate  in a natural language form \cite{SongWHZG18,Wang:qaverif,Tan2018SNetFA} answers to open-domain questions.
Most open-domain QA  approaches adopt the "Retriever Reader" framework: the retriever ranks candidate passages, then the reader constructs the response based on the most relevant retrieved candidate \cite{Min2019MultihopRC,tu-etal-2019-multi,Zhang21}.
Compared with open-domain QA, 
answer generation to open-ended queries has two main specific issues: 
1) all the documents provided by the reader potentially  contribute both as evidence and source to generate the answer leading to difficulties in discriminating between relevance and salience of the spans; 2) while most QA problems target a single-span answer  \cite{yang-etal-2018-hotpotqa} included in one  document, open-ended queries are characterized by multiple facets  \cite{white2009exploratory,Wildemuth12} that could target a multiple-span answer.\\
\indent Our objective is to generate an answer that covers the multiple facets of an open-ended query using as input, an initial ranked list of documents. We basically assume that the list of documents cover the different query facets. A naive approach would be to exploit text-to-text models \cite{JMLR:v21:20-074,Radford2018ImprovingLU}. However, we believe that answering multi-faceted queries would require the modelling of structure prior to generating the answer's content\footnote{As outlined in the Swirl report \cite{Culpepper18},   regarding the importance of structure in the "Generation of Information Objects" (GIOs)}. 
To fit with this requirement, we  adopt a data-to-text (DTT) generation approach \cite{data2019puduppully} that introduces the notion of  structure by guiding the generation with an intermediary plan aiming at determining \textit{what to say} on the basis of the input data.  
This intermediary step, called  \textit{content selection/planning}, reinforces the factualness and the coverage of the generated text since: 1) it  organizes the data structure in a latent form to better fit with the generated output, and 2) it provides a structure to the generated answer based on the elements of the initial  data. Figure \ref{fig:Example} presents an example of a  query from TREC Complex Answer Retrieval (CAR) dataset \cite{dietz2017trec} and the two variants of answers (\textit{plain answers}, \textit{structured answers}) generated by our proposed model trained respectively on two different train-test datasets (See Section 3). 
To sum up, given a ranked list of  documents relevant to a complex information need, this work investigates the potential of content selection and planning DTT generation models for single-turn answer generation.

\vspace{-0.4cm}
\section{A data-to-text approach for answer generation}
\vspace{-0.4cm}
In this section, we introduce the model used for generating natural language answers to open-ended queries formulated by users while completing complex search tasks. 
The designed model is driven by the intuition that the response should be guided by a structure to cover most of the query facets. This prior is modeled through a hierarchical plan which corresponds to a textual object relating the structure of the response with multiple-level titles (titles, subtitles, etc). \\
\indent More formally, we consider a document collection $\mathcal{D}$ and a set $Q \times A \times P$ of query-answer-plan triplets, where $q \in Q$ refer to queries, answers $a \in A$ the final response in natural language provided to the user and plans $p \in P $ represent the hierarchical structure of answers $a$. All documents $d$, queries $q$, answers $a$ are represented by sets of tokens. For modeling the structure of plans $p$, we use $p=\{h_{1}, ..., h_i, ..., h_{|p|}\}$  where  $h_i$ represents a line in the plan expressing a heading (title, subtitles, etc.). The $h_i$ are modeled as sets of tokens. \\
\indent Given a query $q$ and  a document collection $\mathcal{D}$, our objective is  to generate an answer $a$. To do so, we follow the "Retriever Generator" framework  in which: 1) a ranking model $\mathcal{M}_{ret}$ retrieves a ranked list $\mathcal{D}_{q}$ of documents in response to query $q$, where $\mathcal{D}_{q} = \{d_q^1, \dots, d_q^n\}$ and 2) a text generation model $\mathcal{M}_{gen}$ generates  answer $a$ given the retrieved list $\mathcal{D}_{q}$ and query $q$. As outlined earlier, the challenges of our task mainly rely on aggregating information over the ranked list of documents and generating a structured answer in natural language. Thus, we fix the retrieval model $\mathcal{M}_{ret}$ and focus on the generation model $\mathcal{M}_{gen}$. The latter exploits  the DTT generation model based on content selection and  planning \cite{data2019puduppully}. 
 To generate the intermediary plan $p$ and the answer $a$, we rely on two successive encoder-decoders (based on T5 \cite{JMLR:v21:20-074} as the building-box model):\\
$\bullet$ The \textbf{planning encoder-decoder} encodes each document $d_q \in \mathcal{D}_{q}$ concatenated with the query $q$ and decodes a plan $p$. The training of such network is guided by the auto-regressive generation loss:
\begingroup
\setlength\abovedisplayskip{0pt}
\setlength\belowdisplayskip{0pt}
\begin{equation}
    \mathcal{L}_{planning}(q,p)= P(p|q,\mathcal{D}_q) =
    \prod_{j=1}^{|p|} \prod_{k=1}^{|h_j|} P(h_{jk}|h_{j,<k}, q, \mathcal{D}_q)
\end{equation}
\endgroup
where $j$ and $k$ point out resp. to the heading $h_j$ and the $k^{th}$ token $h_{jk}$ in heading $h_j$. $h_{j,<k}$ corresponds to the token sequence in heading $h_j$ before the $k^{th}$ token.\\
$\bullet$ The \textbf{content generation encoder-decoder} encodes each heading $h_p$ in the plan $p$ (generated by the planning encoder-decoder) concatenated with the embedding of the document list $\mathcal{D}_q$. The latter is obtained by the planning encoder-decoder since the T5 model provides embeddings for both documents independently and the set of documents. After the encoding, the network then  decodes an answer $a$. The training  is also guided by the auto-regressive generation loss:
\begingroup
\setlength\abovedisplayskip{0pt}
\setlength\belowdisplayskip{0pt}
\begin{equation}
    \mathcal{L}_{answer}(q,a,p)= P(a|q,p,\mathcal{D}_q) =
    \prod_{k=1}^{|a|} P(a_k|a_{<k},q, p, \mathcal{D}_q)
\end{equation}
\endgroup
where $a_k$ and $a_{<k}$ resp. express the $k^{th}$ token in answer $a$ and the token sequence of answer $a$ before the $k^{th}$ token.\\
\indent The final loss is then a combination of both auto-regressive losses: 
\begingroup
\setlength\abovedisplayskip{0pt}
\setlength\belowdisplayskip{0pt}
\begin{equation}
    \mathcal{L} = \sum_{\{q,a,p\} \in Q \times A \times P} \mathcal{L}_{planning}(q,p) + \mathcal{L}_{answer}(q,a,p)
\end{equation}
\endgroup

\vspace{-0.2cm}
\section{Evaluation Setup}

\vspace{-0.3cm}
\paragraph{Dataset.}
We selected the TREC CAR (Complex Answer Retrieval) 2017 corpus \cite{dietz2017trec}. This dataset includes: (1) queries - denoting complex search tasks with multiple facets, (2) plans - expressing the different expected facets, and (3) paragraphs extracted from English Wikipedia - corresponding to texts associated with plan sections. The TREC CAR task consists of retrieving the paragraphs associated to each plan section to build a structured answer combining both plan sections and paragraphs. We used these \textit{structured answers} as the final objective of our generation model given the queries; and the plans as the structure prior. Due to the structure prior constraint, we removed in the training set\footnote{The large train set was used for training, and the Y1 benchmark test set for testing.} answers without any plans.
To compare the models abilities to generate \textit{structured answers}, we also evaluate
a new form of expected answer (\textit{plain answers}) where structure is not taken into account. For this aim, we built a new dataset upon the initial TREC CAR dataset but only considering the paragraphs (without plans).  Thus, we obtain two versions of datasets (for \textit{structured answers} and \textit{plain answers}) which both  follow the original split of the TREC CAR dataset. Second, for computational reasons, we reduced the number of entries in our training set by considering only a half of Fold 0 of the original dataset. Also, due to the memory constraints of generation models and the length of Wikipedia articles, we reduced the document size by only keeping the first sentence of paragraphs. 
Some statistics of the original  dataset   and our two adapted datasets are given in Table~\ref{datasets3}.

\begin{table}[b]
 \centering
 \scriptsize
 \vspace{-0.4cm}
 \caption{Statistics on the TREC CAR 2017 dataset and its adaptation for experiments.}
 \vspace{-0.2cm}
\begin{tabular}{ p{2.5cm}p{1.5cm}p{1.5cm}p{1.5cm}p{1.5cm} p{1.5cm}p{1.5cm}}
 \hline
 & \multicolumn{2}{c}{Original dataset} & \multicolumn{2}{c}{Structured answers}& \multicolumn{2}{c}{Plain answers}\\
   \cmidrule(r){2-3} \cmidrule(r){4-5} \cmidrule(r){6-7}
  & Train        & Test & Train  & Test & Train  & Test  \\

 \hline
\#answers & 598 308 & 132 &  46 224 & 132  & 46 224 & 132\\
\#tokens/answers &  1376.48 &  5456.94 &  609.31 & 1724.63 & 449.21 & 1409.79\\
\#headings/plan & 6.10& 17.69 & 6.22 & 17.69  & - & -\\
\hline
\end{tabular}
\label{datasets3}
\vspace{-0.7cm}
\end{table}


\vspace{-0.2cm}
\paragraph{Model variants and baselines.}
\label{sec:models}
We implement two versions of our model\footnote{code available at \url{https://github.com/hanane-djeddal/Complex-Answer-Generation/}}: \\
$\bullet$ \textbf{Planning-seq}: a sequential model where the planning module (Equation 1) and the content generation module (Equation 2) are trained separately. At inference, both modules are used sequentially.  \\ 
$\bullet$ \textbf{Planning-e2e}: the end-to-end version of our model (Equation 3). The content module is fed with the output embeddings of the planning module, and document tokens. 

Besides, we compare our models with two baselines:  1) the  \textbf{T5} model \cite{JMLR:v21:20-074} which is fine-tuned on each dataset, 
and 2) \textbf{Ext}, an extractive method  where we extract, for each sentence in the ground truth, a sentence in the input supporting documents that maximizes the F1 score of BERTScore \cite{zhang2019bertscore}. 
All models consider for each topic a set of 10 relevant paragraphs ranked using BM25 as input.

\paragraph{Metrics.}
To evaluate the quality of the generation, we consider three well-known metrics: 1) the ROUGE-L mid
metric (Rouge-P, Rouge-R, Rouge-F) \cite{lin2004rouge} measuring the exact match between the generated and the reference texts, 2) the BERTScore \cite{zhang2019bertscore} (the F1 score is reported) which computes similarity between the generated and the gold reference text  embeddings, 3) the QuestEval \cite{scialom2021questeval} framework which relies on question answering models to assess whether a summary contains all the source information: if the same questions are asked to the generated and the reference texts, the produced answers should be consistent.


To evaluate the model's ability to generate structure (namely the plans), we use  the METEOR score \cite{banerjee2005meteor}  capturing how well-ordered the output words are. 


\vspace{-0.3cm}
\section{Results}
\vspace{-0.3cm}

\begin{table}[t]
    \centering
    \caption{Effectiveness of the answer generation. In bold are the highest metric value among the generation models (T5, Planning-seq, Planning-e2e).} 
    \vspace{-0.2cm}
    \begin{tabular}{ccccccccc}
    \hline
      &&& \ttitle{\# tokens}&\ttitle{Rouge-P} & \ttitle{Rouge-R} & \ttitle{Rouge-F} & \ttitle{BERTScore} & \ttitle{QuestEval} \\
      \hline   
       \parbox[t]{3.0mm}{\multirow{4}{*}{\rotatebox[origin=c]{90}{\ttitle{\begin{scriptsize}structured \end{scriptsize}}}}} & \parbox[t]{3.0mm}{\multirow{4}{*}{\rotatebox[origin=c]{90}{\ttitle{\begin{scriptsize} answers\end{scriptsize}}}}} &  EXT &$898.22$&$36.50$ &  $26.99$  & $29.86$  &  $85.50$ & $41.99$\\
       & & T5 &$126.25$& $\boldsymbol{76.19}$ & $08.41$ & $14.25$ & $\boldsymbol{84.95}$  &  $39.06$  \\
       & & Planning-seq&$181.39$& $62.94$ &$09.57$&$ 15.36$ &$84.44$& $37.47$\\
       & & Planning-e2e &$203.48$& $63.4$ & $\boldsymbol{10.21}$ & $\boldsymbol{16.09}$ & $84.91$ &  $\boldsymbol{39.31}$ \\
        \hline
       \parbox[t]{3.0mm}{\multirow{4}{*}{\rotatebox[origin=c]{90}{\ttitle{\begin{scriptsize}plain  \end{scriptsize}}}}} & \parbox[t]{3.0mm}{\multirow{4}{*}{\rotatebox[origin=c]{90}{\ttitle{\begin{scriptsize}  answers\end{scriptsize}}}}}& EXT &885.35&  $34.35$ &  $26.73$  & $28.99$  &  $86.30$ & $42.34$ \\
        && T5 & $110.62$& $\boldsymbol{78.05}$& $09.24$& $15.48$& $85.51$& $ 39.89$ \\
        && Planning-seq&$163.58$& $65.73$& $ 10.34$ &$16.27$ &$84.29 $ & $38.46$\\
        && Planning-e2e & $126.91$& $75.92$ & $\boldsymbol{10.34}$& $\boldsymbol{17.05}$ &$\boldsymbol{85.67}$ &$\boldsymbol{40.78}$\\ 
        \hline
    \end{tabular}
    \label{tab:fold0_results}
    \vspace{-0.4cm}
\end{table}

We perform the experimental evaluations w.r.t. two objectives. First, we measure the effectiveness of the generated answers. Second, we provide a thorough analysis of the generated plans.  
 \vspace{-0.3cm}
\paragraph{Answer generation effectiveness.}
Table \ref{tab:fold0_results} reports the results of the different settings and models used for generating answers.  
It is worth of recall that the EXT baseline does not address the generation task and is built from the ground truth leading to provide high value trends. 
With this in mind, we can outline  the following statements:\\
$\bullet$ Planning-based generation models are competitive regarding the T5 generation baseline: our models allow to generate longer answers (avg. 200 tokens), thus increasing the recall metric (Rouge-R). The smaller precision (76.19 for T5, up to 63.4 for our models) does not hinder the semantic content of the answer (see BERTScore and QuestEval values which are very close to the EXT metrics). This suggests that our models are able to generate answers with the adequate content, even if noisy at some points. \\
$\bullet$ One can see the general trend towards higher metrics for all models in the \textit{plain answers} setting compared to the \textit{structured answers} setting (e.g. Rouge-P reaching up to 78.05 vs. 76.15) over all models. The \textit{plain answers} setting is less difficult since the expected answer is not structured (only composed of paragraphs); evaluation metrics are higher since the gold reference is not based on  both plans and paragraphs (as in the \textit{structured answers} setting). In the \textit{plain answers} setting,  our models are most effective (with an advantage for Planning-e2e with, for instance 17.05 Rouge-F vs. 15.48 for T5). Even if the  \textit{plain answers} setting does not expect plans in the final answer, our models generate an intermediary plan that guides the answer generation. In contrast,   T5  directly generates the answer. This reinforces our intuition about the importance of structure prior for generating an answer to a complex information need. \\
$\bullet$ Our end-to-end model seems more effective than the sequential one (e.g., resp. 40.78 vs. 38.46 for the QuestEval metric), suggesting the relevance of guiding the learning of the planning encoder-decoder by the answer generation task.

\begin{table}[t]
    \centering
    \scriptsize
    \caption{Analysis of the intermediate and final plans (resp. noted IP and FP) in our sequential and end-to-end planning-based models for the \textit{structured answers} setting. }
    \vspace{-0.2cm}
    \begin{tabular}{lcccccccccc}
     \hline
      \multicolumn{2}{c}{} & \ttitle{\#tokens}& \ttitle{\#heading}& \ttitle{depth}&\ttitle{Rouge-P} & \ttitle{Rouge-R} & \ttitle{Rouge-F} & \ttitle{BERTScore} & \ttitle{Meteor} \\
      \hline   
      T5 & FP& $1.41$ & $2.24$ &$1.14$& $\boldsymbol{39.89}$&$04.69$ &$07.69$& $77.40$ & $3.24$ \\
      \hdashline
      \multirow{2}{*}{Planning-seq}& IP&$1.83$&$\boldsymbol{4.42}$&$\boldsymbol{1.45}$& $31.20$  & $\boldsymbol{8.29}$ &$\boldsymbol{11.51}$& $\boldsymbol{81.25}$&$\boldsymbol{5.97}$\\
           & FP &$\boldsymbol{1.88}$&$4.11$&$\boldsymbol{1.45}$&$31.31$ & $7.93$& $11.03 $ & $80.49$ &$5.55$ \\
           \hdashline
       \multirow{2}{*}{Planning-e2e}&  IP &$1.57$ &$3.37$ &$1.15$& $35.15$ & $07.34$ & $11.12$ & $81.21$ & $5.51$ \\
        & FP & $1.64$& $3.27$ &$1.16$& $34.79$ & $06.38 $&$09.78 $ & $ 80.70$ & $ 4.71$  \\

        \bottomrule
    \end{tabular}
    \label{tab:my_label}
    \vspace{-0.4cm}
\end{table}

 \vspace{-0.3cm}
\paragraph{Analyses of the generated plans.}
To get a deeper understanding of our model behavior regarding the structure prior,  we analyze the plans generated  by the different encoder-decoders: the intermediate one provided by the planning encoder-decoder  and the final one included in the final answer after the generation encoder-decoder (we simply extracted headings of the structured answer - red and blue lines in Figure 1).
We report in Table \ref{tab:my_label} the different evaluation metrics presented in Section 4 to measure the quality of plans and add some plan statistics (the average number of tokens for each plan section - \#token; the number of generated plan sections by query -\#heading; the mean depth of plan sections i.e $i$ of $h_i$ -\textit{depth}). 
Comparison of intermediate and final plans obtained by our models with the final one generated by T5 highlights that: 1) our plans are longer and more complex (more tokens by plan section - up to 1.83 in average, more and deeper headings - up to 4/5 headings in average), 2) our plans generally cover more facets (higher recall), in correct order (higher Meteor) with a better relevant semantics (higher BERTScore). The lowest precision (up to 35.15 vs. 39.89 for the T5) might be explained by the plan sizes.
Moreover, the comparison of intermediate vs. final plans underlines a general trend towards lower quality of plans in the final step (e.g., 11.51 vs. 11.03 for Planning-seq in terms of  Rouge-F). But the previous discussion on answer effectiveness, 
and the higher performance of our models regarding T5) suggests that there is a balance to reach between raw text and plan generation and that the structure prior is however highly beneficial for generating a good answer.



\vspace{-0.4 cm}
\section{Conclusion}
\vspace{-0.2 cm}
Traditionally, IR approaches solving complex information needs focused on leveraging  multi-turn interactions to provide optimal rankings of candidate documents at each turn. In this paper we have suggested  alternative retrieval models that do not rely on the interactive updating of queries and document rankings as answers. We suggest one such alternative approach can be found using data-to-text generation models to generate in a single-turn, a natural language and structured answer. Experimental evaluation of a planning-based  DTT model using the TREC CAR dataset shows the potential of our intuition.  We believe that our work opens up novel areas of investigation including answer generation and explanation in conversational systems for IR.

\section*{Acknowledgement}
We would like to thank projects ANR JCJC SESAMS (ANR-18- CE23-0001) and ANR COST (ANR-18-CE23-0016) for supporting this work. This work was performed using HPC resources from GENCI-IDRIS (Grant 2021-101681).

\bibliographystyle{spmpsci}
\bibliography{bibliography}

\end{document}